\documentclass[10pt,journal,compsoc]{IEEEtran}

\usepackage{epsfig}
\usepackage{graphicx}
\usepackage{amsmath}
\usepackage{amssymb}
\usepackage{mathtools}
\usepackage{bbm}
\usepackage{makecell, verbatim, sidecap}

\usepackage[nocompress]{cite}

\usepackage{balance}

\usepackage[breaklinks=true,bookmarks=false]{hyperref}

\usepackage{color}

\def\R{{\mathbb R}}

\DeclarePairedDelimiter{\floor}{\lfloor}{\rfloor}

\def\centerness{ {  \rm centerness  } }

\renewcommand{\texttt}[1]{${{\tt #1}}$}

\usepackage{xspace}

\renewcommand{\vec}[1]{\ensuremath{\pmb{#1}}}
\newcommand{\mat}[1]{\ensuremath{\mathbf{#1}}}
\newcommand{\set}[1]{\ensuremath{\mathscr{#1}}}

\makeatletter
\@tfor\next:=abcdefghijklmnopqrstuvwxyxz\do
{\begingroup\edef\x{\endgroup
        \noexpand\@namedef{v\next}{\noexpand\vec{\next}}%
    }\x}
\@tfor\next:=ABCDEFGHIJKLMNOPQRSTUVWXYZ\do
{\begingroup\edef\x{\endgroup
        \noexpand\@namedef{m\next}{\noexpand\mat{\next}}%
    }\x}
\@tfor\next:=ABCDEFGHIJKLMNOPQRSTUVWXYZ\do
{\begingroup\edef\x{\endgroup
        \noexpand\@namedef{s\next}{\noexpand\set{\next}}%
    }\x}
\makeatother

\def\Ours{{FCOS}\xspace}
\def\eg{{\it e.g.}\xspace}
\def\ie{{\it i.e.}\xspace}

\renewcommand{\paragraph}{\textbf}

\begin{document}

    \title{\Ours: A Simple and Strong Anchor-free\\ Object Detector}

    \author{
         Zhi Tian,
     Chunhua Shen,
         Hao Chen,
          Tong He
        \thanks{
            Authors are with
            The University of Adelaide, Australia.
            C. Shen is the
            corresponding author  (e-mail: chunhua.shen@icloud.com).

            Accepted to
            IEEE Transactions on Pattern Analysis and Machine Intelligence.
        }
    }

    \IEEEtitleabstractindextext{

        \begin{abstract}

            In computer vision, object detection is one of most important tasks,
            which underpins a few instance-level recognition tasks and many downstream applications.
            Recently one-stage methods have gained much attention over two-stage approaches due to their  simpler design and competitive performance.
            Here we propose a fully convolutional one-stage object detector (FCOS) to solve object detection in a per-pixel prediction fashion, analogue to
            other dense prediction problems such as
            semantic segmentation. Almost all state-of-the-art object detectors such as RetinaNet, SSD,  YOLOv3, and Faster R-CNN rely on pre-defined anchor boxes. In contrast, our proposed detector FCOS is anchor box free, as well as proposal free. By eliminating the pre-defined set of anchor boxes, FCOS completely avoids the complicated computation related to anchor boxes such as calculating %
            the intersection over union (IoU) scores
            during training. More importantly, we also avoid all hyper-parameters related to anchor boxes, which are often
            sensitive to the final detection performance.
            With the only post-processing non-maximum suppression (NMS),
            we demonstrate a much simpler and flexible detection framework achieving improved detection accuracy. We hope that the proposed FCOS framework can serve as a simple and strong alternative for many other instance-level tasks.
            Code is available at:
            \href{https://git.io/AdelaiDet}{{ $\tt git.io/AdelaiDet$}}

        \end{abstract}

        \begin{IEEEkeywords}
            Object detection, fully convolutional one-stage object detection,
            anchor box,
            deep learning.
        \end{IEEEkeywords}
    }

    \maketitle

    \section{Introduction}
    \begin{figure}[t!]
        \centering
        \includegraphics[width=\linewidth]{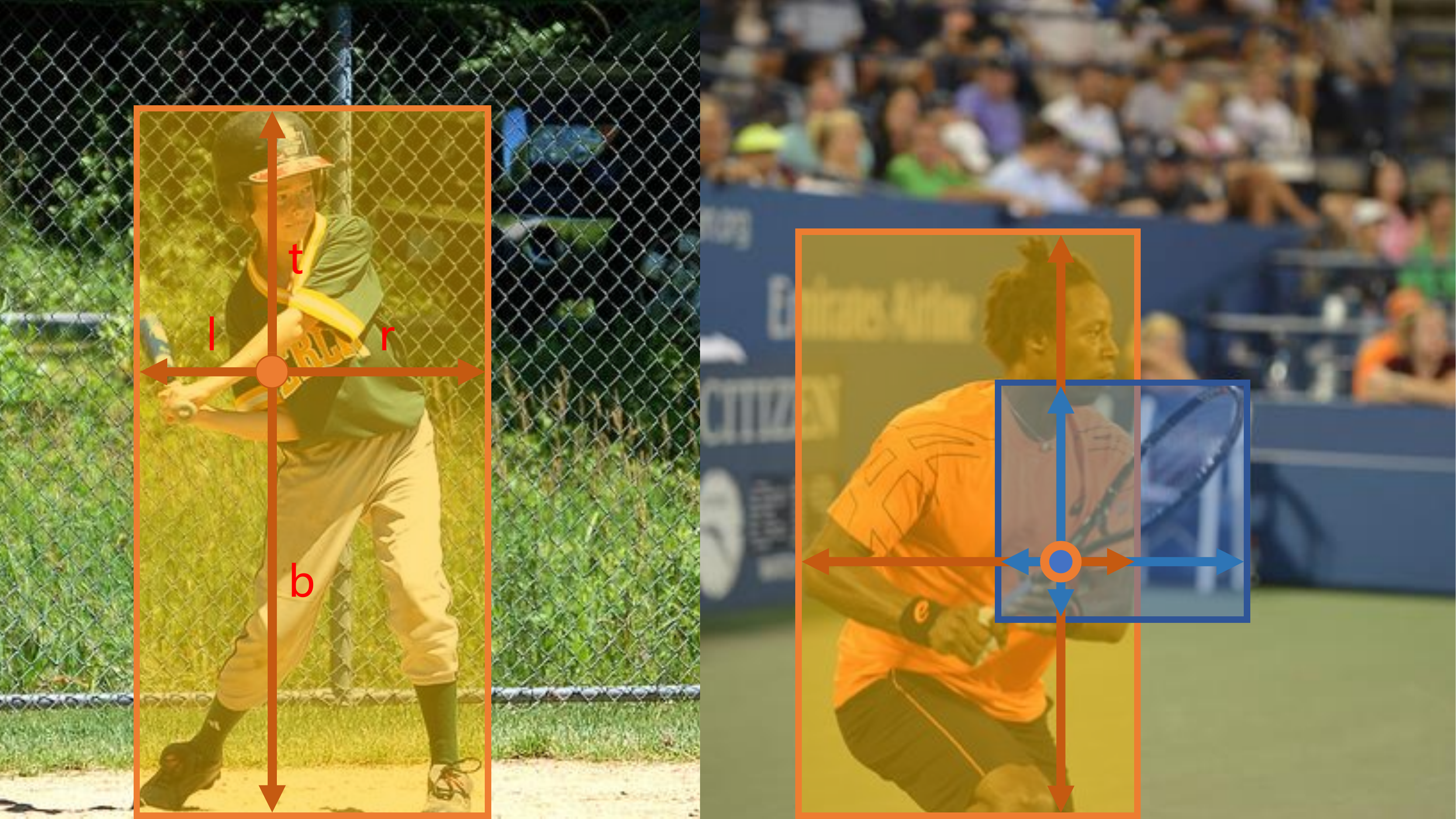}
        \caption{
        \textbf{Overall concept of \Ours{}}.
        As shown in the left image, \Ours\ works by predicting a 4D vector $(l, t, r, b)$ encoding the location of a bounding box at each foreground pixel (supervised by ground-truth bounding box information during training).
            The right plot shows that when a location residing in multiple bounding boxes, it can be ambiguous in terms of which bounding box this location should regress.
        }
        \label{fig:training_targets}
    \end{figure}

    Object detection
    requires %
    an
    algorithm to predict a bounding box location
    and %
    a category label for each instance of interest in an image.
    Prior to deep learning, the sliding-window approach was the main method
    \cite{Viola01robust,FisherBoost2013IJCV,DollarPAMI14pyramids}, which exhaustively classifies every possible location, thus requiring feature extraction and
    classification evaluation to be very fast. With deep learning,  detection has been largely shifted to the use of fully convolutional networks (FCNs) since the invention of
    Faster R-CNN \cite{ren2015faster}.
    All current mainstream detectors such as Faster R-CNN \cite{ren2015faster}, SSD \cite{liu2016ssd} and YOLOv2, v3 \cite{redmon2018yolov3} rely on a set of pre-defined anchor boxes and {\it it has long been believed  that
        the use of anchor boxes is  the key to  modern
        detectors' success.} Despite their great success, it is important to note that anchor-based detectors suffer some drawbacks:
        \begin{itemize}
            \item
    As shown in Faster R-CNN and RetinaNet \cite{lin2017focal},
    detection performance is sensitive to
    the sizes, aspect ratios and number of anchor boxes.
    For example, in RetinaNet, varying these hyper-parameters affects the performance up to $4\%$ in AP on the COCO benchmark \cite{lin2014microsoft}. As a result, these hyper-parameters need to be carefully tuned  in  anchor-based detectors.
    \item  Even with careful design, because
    the scales and aspect ratios of anchor boxes are kept fixed,
    detectors encounter difficulties to deal with object candidates with large shape variations, particularly for small objects. The pre-defined anchor boxes also hamper
    the generalization ability of detectors, as they need to be re-designed on new detection tasks with different object sizes or aspect ratios.
    \item  In order to achieve a high recall rate, an anchor-based detector is required to densely place anchor boxes on the input image (\eg, more than 180K anchor boxes in feature pyramid networks (FPN) \cite{lin2017feature} for an image with its shorter side being 800). Most of these anchor boxes are labeled as negative samples during training. The excessive  number of negative samples aggravates the imbalance between positive and negative samples in training.
    \item
    Anchor boxes also involve complicated computation such as calculating the intersection-over-union (IoU) scores with ground-truth bounding boxes.

    \end{itemize}

    Recently,
    per-pixel prediction FCNs \cite{long2015fully} have achieved tremendous success in dense prediction tasks such as semantic segmentation \cite{long2015fully, tian2019decoders, He_2019_CVPR}, depth estimation \cite{Depth2015Liu, Yin2019enforcing}, keypoint detection \cite{ICCV2017Chen} and counting.
    As one of high-level vision tasks, object detection might be the only one deviating from the neat fully convolutional per-pixel prediction framework mainly due to the use of anchor boxes.

    It is
    natural
    to ask a question: {\it Can we solve object detection in the neat per-pixel prediction fashion, analogue to FCN for semantic segmentation, for example?
    } Thus those fundamental vision tasks can be unified in (almost) one single framework.
    We show in this work  that the answer is affirmative.
    Moreover, we demonstrate that, the much simpler FCN-based detector can surprisingly achieve even better performance than its anchor-based counterparts.

    In the literature, some works attempted to leverage the per-pixel prediction FCNs for object detection such as DenseBox \cite{huang2015densebox}. Specifically, these FCN-based frameworks directly predict a 4D vector plus a class category at each spatial location on a level of feature maps. As shown in  Fig.~\ref{fig:training_targets} (left), the 4D vector depicts the relative offsets from the four sides of a bounding box to the location. These frameworks are  similar to the FCNs for semantic segmentation, except that each location is required to regress a 4D continuous vector.

    However, to handle the bounding boxes with different sizes, DenseBox \cite{huang2015densebox} crops and resizes training images to a fixed scale. Thus DenseBox has to perform detection on image pyramids,
    which is against FCN's philosophy of computing all convolutions once.

    Besides, more significantly,
    these methods are mainly used in special domain objection detection such as scene text detection \cite{zhou2017east, he2018end} or face detection \cite{yu2016unitbox, huang2015densebox},
    since it is believed that these methods do not work well when applied to generic object detection with highly overlapped bounding boxes. As shown in  Fig.~\ref{fig:training_targets} (right), the highly overlapped
    bounding boxes result in an intractable ambiguity: it is not clear w.r.t.\ which bounding box to regress for the pixels in the overlapped regions.

\begin{figure*}[ht!]
        \centering
        \includegraphics[width=.90099895\linewidth]{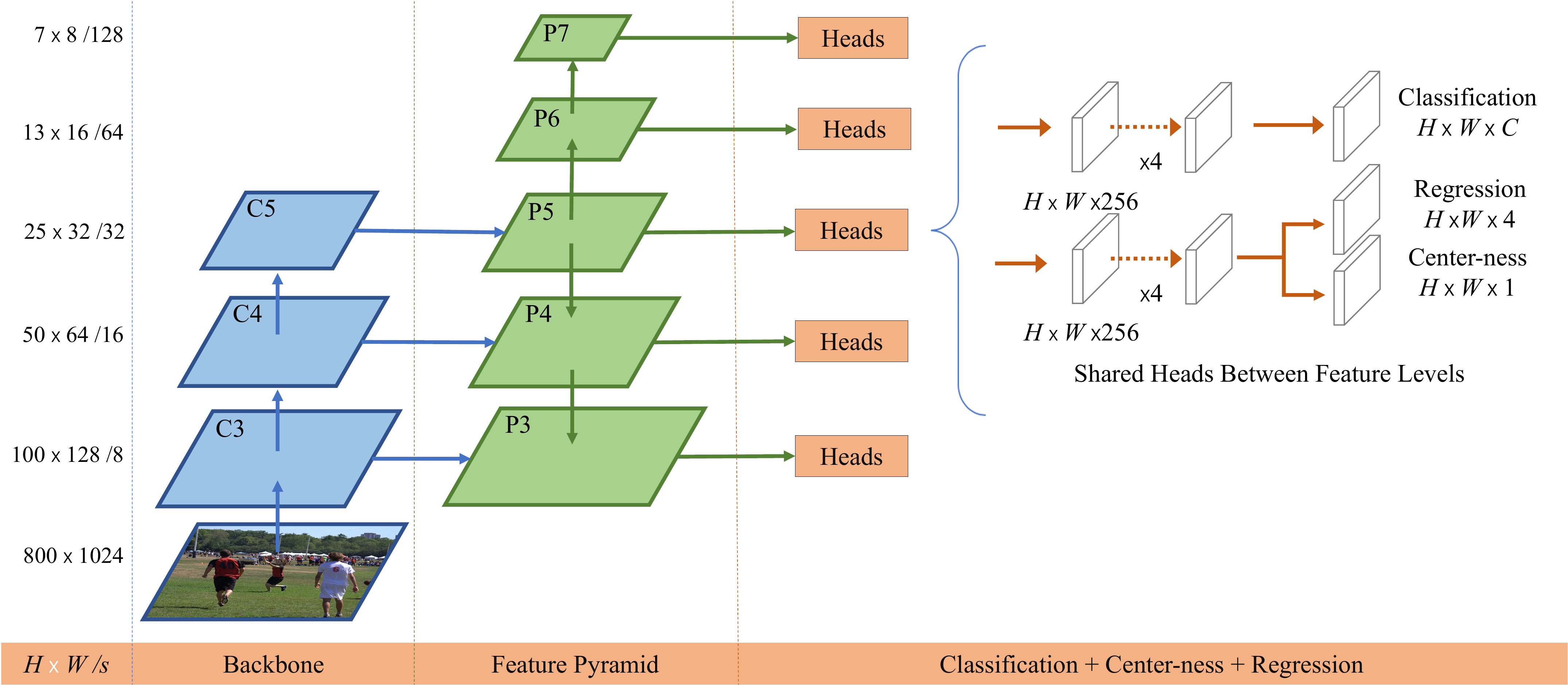}
        \caption{\textbf{The network architecture of \Ours}, where C3, C4, and C5 denote the feature maps of the backbone network and P3 to P7 are the feature levels used for the final prediction.
            $H \times W$ is the height and width of feature maps.
            `/$s$' ($s=8, 16, ..., 128$) is the down-sampling ratio of the feature maps at the level to the input image.
            As an example, all the numbers are computed with an $ 800\times 1024$ input.
        }
        \label{fig:main_figure}
    \end{figure*}

    In the sequel, we take a closer look at the issue and show that with FPN this ambiguity can be largely eliminated. As a result, our method can already obtain similar or even better detection accuracy with those traditional anchor based detectors. Furthermore, we observe that our method may produce a number of low-quality predicted bounding boxes at the locations that are far from the center of an target object. It is easy to see that the locations near the center of its target bounding box can make more reliable predictions. As a result, we introduce a novel ``center-ness'' score to depict the deviation of a location to the center, as defined in Eq.~\eqref{eq:centerness}, which is used to down-weigh low-quality detected bounding boxes and thus helps to suppress these low-quality detections in NMS. The center-ness score is predicted by a branch (only one layer) in parallel with the bounding box regression branch, as shown in Fig.~\ref{fig:main_figure}. The simple yet effective center-ness branch remarkably improves the detection performance with a negligible increase in computational time.

    This new detection framework enjoys the following advantages.
    \begin{itemize}

        \item Detection is now unified with many other FCN-solvable tasks such as semantic segmentation, making it easier to re-use ideas from those tasks.
        An example is shown in \cite{Liu2020TPAMI}, where a structured knowledge distillation method was developed for dense prediction tasks.
        Thanks to the standard FCN framework of FCOS,
        the developed technique can be immediately applied to FCOS based
        object detection.

        \item Detection becomes proposal free and anchor free, which significantly reduces the number of design parameters. The design parameters typically need heuristic tuning and many tricks are involved in order to achieve good performance. Therefore, our new detection framework makes the detector, particularly its training, \emph{considerably} simpler.
        \item By eliminating the anchor boxes, our new detector completely avoids the complicated computation related to anchor boxes such as the IOU computation and matching between the anchor boxes and ground-truth boxes during training, resulting in faster training and testing than its anchor-based counterpart.
        \item Without bells and whistles, we achieve state-of-the-art results among one-stage detectors. Given
        its improved
        accuracy
        of the much simpler anchor-free detector,  {\it we encourage the community to rethink the necessity of anchor boxes in object detection}, which are currently considered as the \emph{de facto} standard for designing  detection methods.

        \item With considerably reduced design complexity, our proposed detector outperforms previous strong baseline detectors such as Faster R-CNN \cite{ren2015faster}, RetinaNet \cite{lin2017focal}, YOLOv3 \cite{redmon2018yolov3} and SSD \cite{liu2016ssd}.
        More importantly,  due to its simple design,
        FCOS can be easily extended to solve other instance-level recognition tasks with minimal modification, as already evidenced by instance segmentation
        \cite{PolarMask,MEInst,CenterMask,chen2020blendmask}, keypoint detection \cite{tian2019directpose},
        text spotting \cite{ABCNet}, and tracking \cite{tracking2020Wang,tracking2020Guo}.
        We expect to see more instance recognition methods built upon FCOS.

    \end{itemize}

    \subsection{Related Work}
    Here we review some work that is closest to ours.

    \paragraph{Anchor-based Detectors.} Anchor-based detectors inherit the ideas from traditional sliding-window  and
    proposal based detectors such as Fast R-CNN \cite{girshick2015fast}.
    In anchor-based detectors, the anchor boxes can be viewed as pre-defined sliding windows or proposals, which are classified as positive or negative patches, with an extra offsets regression to refine the prediction of bounding box locations. Therefore, the anchor boxes in these detectors may be viewed as \emph{training samples}. Unlike previous detectors like Fast RCNN, which compute image features for each sliding window/proposal repeatedly,
    anchor boxes  make use of the feature maps of CNNs and avoid
    repeated feature computation, speeding up detection process dramatically. The design of anchor boxes are popularized by Faster R-CNN in its RPNs \cite{ren2015faster}, SSD
    \cite{liu2016ssd} and YOLOv2 \cite{redmon2017yolo9000},
    and has become the %
    convention
    in a modern detector.

    However, as described above, anchor boxes result in excessively  many hyper-parameters, which typically
    need to be carefully tuned   in order to achieve good performance. Besides the above hyper-parameters describing anchor shapes, the anchor-based detectors also need other hyper-parameters to label each anchor box as a positive, ignored or negative sample.
    In previous works, they often employ intersection over union (IOU) between anchor boxes and ground-truth boxes to determine the label of an anchor box (\eg, a positive anchor if its IOU is in $[0.5, 1]$). These hyper-parameters have shown a great impact on the final accuracy, and require heuristic tuning. Meanwhile, these hyper-parameters are specific to detection tasks, making detection tasks deviate from a neat fully convolutional network architectures used in other dense prediction tasks such as semantic segmentation.

    \paragraph{Anchor-free Detectors.} The most popular anchor-free detector might be
    YOLOv1 \cite{redmon2016you}. Instead of using anchor boxes, YOLOv1 predicts bounding boxes at points near the center of objects. Only the points near the center are used since they are considered to be able to produce higher-quality detection. However, since only points near the center are used to predict bounding boxes, YOLOv1 suffers from low recall as mentioned in YOLOv2 \cite{redmon2017yolo9000}. As a result, YOLOv2 \cite{redmon2017yolo9000} employs anchor boxes as well. Compared to YOLOv1, \Ours\ can take advantages of all points in a ground truth bounding box to predict the bounding boxes and the low-quality detected bounding boxes can be suppressed by the proposed ``center-ness" branch.  As a result, \Ours\ is able to provide comparable recall with anchor-based detectors as shown in our experiments.

    CornerNet \cite{law2018cornernet} is a recently proposed one-stage anchor-free detector, which detects a pair of corners of a bounding box and groups them to form the final detected bounding box. %
    CornerNet requires much more complicated post-processing to group the pairs of corners belonging to the same instance. An extra distance metric is learned for the purpose of grouping.

    Another family of anchor-free detectors such as \cite{yu2016unitbox} are based on DenseBox~\cite{huang2015densebox}. %
    The family of detectors have been considered  unsuitable for generic object detection due to
    difficulty in handling overlapping bounding boxes and  the recall being relatively low.
    In this work, we show that both problems can be largely alleviated with multi-level prediction (\ie, FPNs). Moreover, we also show together with our proposed center-ness branch, the much simpler detector can achieve much better detection performance than its anchor-based counterparts. Recently, FSAF \cite{zhu2019feature} was proposed to employ an anchor-free detection branch as a complement to an anchor-based detection branch since they consider that a totally anchor-free detector cannot achieve good performance. They also make use of a feature selection module to improve the performance of the anchor-free branch, making the anchor-free detector have a comparable performance to its anchor-based counterpart. However, in this work, we surprisingly show that the totally anchor-free detector can actually obtain better performance than its anchor-based counterpart, without the need for the feature selection module in FSAF. Even more surprisingly, it can outperform the combination of anchor-free and anchor-based detectors in FSAF. As a result, the long-standing anchor-boxes can be completely eliminated,
    making detection
    significantly
    simpler.

    There are also some concurrent anchor-free detectors. For example, CenterNet~\cite{zhou2019objects} predicts the center, width and height of objects with hourglass networks, demonstrating promising performance. Compared to CenterNet, \Ours\ enjoys faster training, and has a better accuracy/speed trade-off, as shown in Table~\ref{table:real_time}. RepPoints~\cite{yang2019reppoints} represents the boxes by a set of points and uses converting functions to get the target boxes. In contrast, \Ours\ is a concise and very straightforward method for box detection.

    After the submission of this work, some new anchor-free detectors %
    appeared.
    For instance, CPN~\cite{duan2020corner} replaces the region proposal networks (RPNs) in two-stage detectors with bottom-up anchor-free networks, achieving
    improved
    results. HoughNet~\cite{samet2020houghnet} employs a voting mechanism to improve the performance of anchor-free bottom-up detection. These detectors require a grouping or voting post-processing, thus being considerably more complicated
    and slower
    than \Ours. There are also many follow-up works built upon \Ours~\cite{qiu2020borderdet, li2020generalized, zhang2020bridging, zhu2020autoassign}. These works enhance the detection features, loss functions or the assigning strategy of \Ours, further boosting the anchor-free detector's performance.

        \begin{figure}[t!]
        \centering
        \includegraphics[width=0.9059\linewidth]{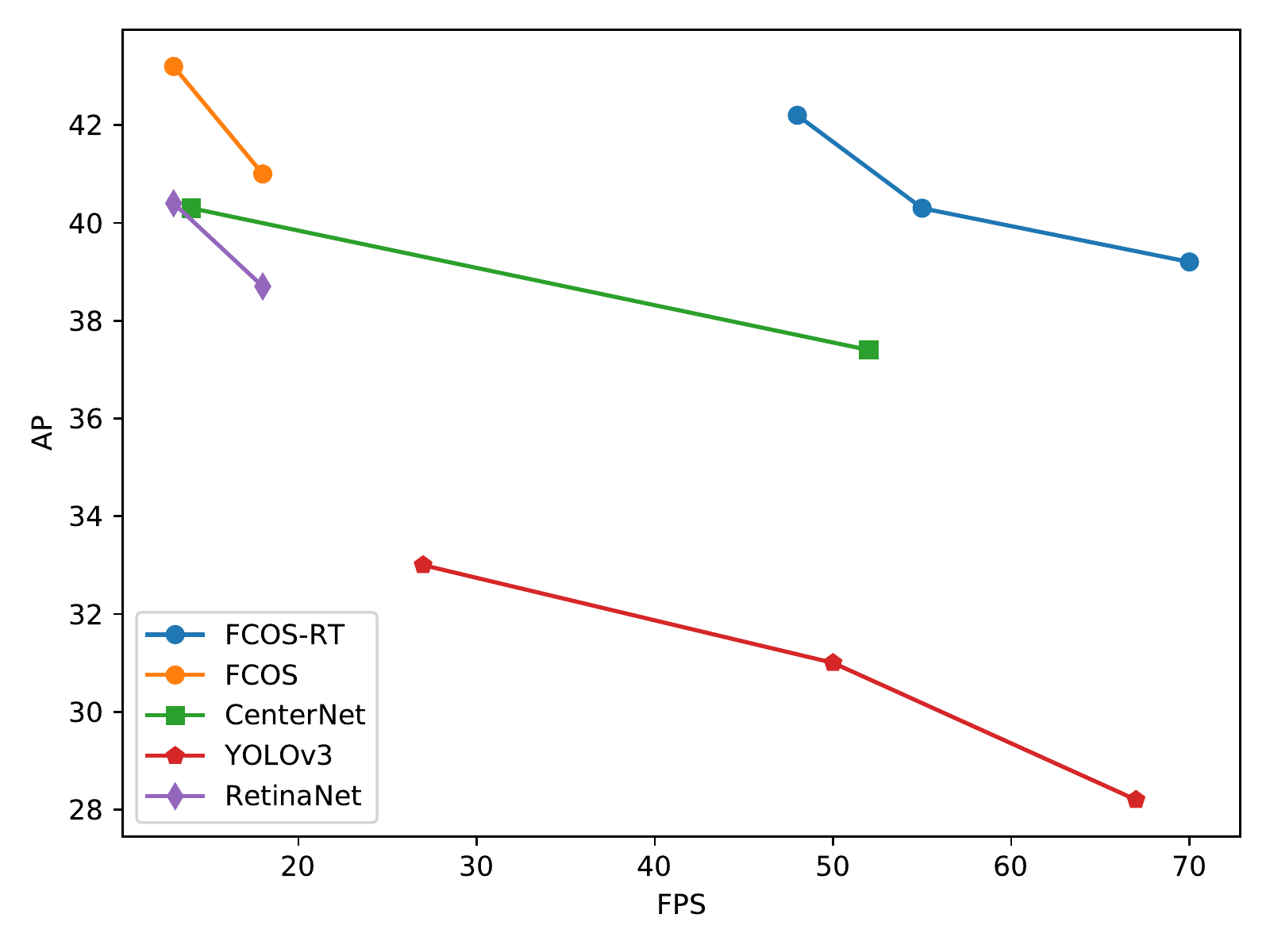}
        \caption{\textbf{Speed/accuracy trade-off between \Ours and
        several recent methods}: CenterNet \cite{zhou2019objects},
        YOLOv3 \cite{redmon2018yolov3} and RetinaNet \cite{lin2017focal}. Speed is measured on a NVIDIA 1080Ti GPU.
        For fair comparison, we only measure the network latency for all detectors. RetinaNet results are from \texttt{Detectron2}.
        \Ours\ achieves competitive performance compared with  recent methods including  anchor-based ones.
        }
        \label{fig:speed_acc}
    \end{figure}

    \section{Our Approach}
    In this section, we first reformulate object detection in a per-pixel prediction fashion. Next, we show that how we make use of multi-level prediction to improve the recall and resolve the ambiguity resulted from overlapped bounding boxes. Finally, we present our proposed ``center-ness" branch, which helps suppress the low-quality detected bounding boxes and improves the overall performance by a large margin.

    \subsection{Fully Convolutional One-Stage Object Detector%
    }\label{sec:main_method}
    Let $F_i \in \R^{H \times W \times C}$ be the feature maps at layer $i$ of a backbone CNN and $s$ be the total stride until the layer. The ground-truth bounding boxes for an input image are defined as $\{B_i\}$, where $B_i = (x^{(i)}_0, y^{(i)}_0, x^{(i)}_1 y^{(i)}_1, c^{(i)}) \in \R^4 \times \{1, 2\ ...\ C\}$.
    Here
    $(x^{(i)}_0, y^{(i)}_0)$ and $(x^{(i)}_1 y^{(i)}_1)$ %
    denote the coordinates of the left-top
    and right-bottom corners of the bounding box.
    $c^{(i)}$ is
    the class that the object in the bounding box belongs to. %
    $C$ is the number of classes, which is $80$ for the MS-COCO dataset.

    For each location $(x, y)$ on the feature map $F_i$, we can map it back onto the input image as $(\floor{\frac{s}{2}} + xs, \floor*{\frac{s}{2}} + ys)$, which is near the center of the receptive field of the location $(x, y)$. Different from anchor-based detectors, which consider the location on the input image as the center of (multiple) anchor boxes and
    regress the target bounding box with these anchor boxes as references, we directly %
    regress the target bounding box at the location. In other words, our detector directly views locations as \emph{training samples} instead of anchor boxes in anchor-based detectors, which is the same as FCNs for semantic segmentation \cite{long2015fully}.

    Specifically, location $(x, y)$ is considered as a positive sample if it falls into the center area of any ground-truth box, by following \cite{fcosplus}. The center area of a box centered at $(c_x, c_y)$ is defined as the sub-box $(c_x - rs, c_y - rs, c_x + rs, c_y + rs)$, where $s$ is the total stride until the current feature maps and $r$ is a hyper-parameter being $1.5$ on COCO. The sub-box is clipped so that it is not beyond the original box. Note that this is different from our original conference version \cite{tian2019fcos}, where we consider the locations positive as long as they are in a ground-truth box. The class label $c^*$ of the location is the class label of the ground-truth box.
    Otherwise it is a negative sample and $c^* = 0$ (background class). Besides the label for classification, we also have a 4D real vector $\vt^* = (l^*, t^*, r^*, b^*)$
    being the regression targets for the location. Here $l^*$, $t^*$, $r^*$ and $b^*$ are the distances from the location to the four sides of the bounding box,
    as shown in Fig.~\ref{fig:training_targets} (left). If a location falls into the center area of multiple bounding boxes, it is considered as an \emph{ambiguous sample}. We simply choose the bounding box with minimal area as its regression target. In the next section, we will show that with multi-level prediction, the number of ambiguous samples can be reduced significantly and thus they hardly affect the detection performance. Formally, if location $(x, y)$ is associated to a bounding box $B_i$, the training regression targets for the location can be formulated as,
    \begin{equation} \label{regression_targets}
    \begin{aligned}
    l^* = (x - x^{(i)}_0) / s,~~t^* = (y - y^{(i)}_0) / s, \\
    r^* = (x^{(i)}_1 - x) / s,~~b^* = (y^{(i)}_1 - y) / s,
    \end{aligned}
    \end{equation}
    where $s$ is the total stride until the feature maps $F_i$, which is used to scale down regression targets and prevents the gradients from exploding during training. Together with these designs, \Ours can detect objects in an anchor-free way and everything is learned by the networks without the need for any pre-defined anchor-boxes. \emph{It is worth noting that this is not identical to an anchor-based detector with one anchor-box per location}, the crucial difference is the way we define positive and negative samples. The single-anchor detector still uses pre-defined anchor-boxes as a prior and uses IoUs between the anchor-boxes and ground-truth boxes to determine the labels for these anchor-boxes. In \Ours, we remove the need for the prior and the locations are labeled by their inclusion in ground-truth boxes. In experiments, we will show that using a single anchor can only achieve inferior performance.

    \paragraph{Network Outputs.} Corresponding to the training targets, the final layer of our networks predicts an 80D vector $\vp$ for classification and a 4D vector  $\vt =(l, t, r, b)$ encoding bounding-box coordinates.
    Following \cite{lin2017focal}, instead of training a multi-class classifier,
    we train $ C $ binary classifiers.
    Similar to \cite{lin2017focal}, we add two branches, respectively with four convolutional layers (exclude the final prediction layers) after the feature maps produced by FPNs for classification and regression tasks, respectively.
    Moreover, since the regression targets are always positive, we employ ${\rm ReLU}(x)$ to map any real number to $(0, \infty)$ on the top of the regression branch. \emph{It is worth noting that \Ours\ has $9\times$ fewer network output variables  than the popular anchor-based detectors \cite{lin2017focal, ren2015faster} with 9 anchor boxes per location,} which is of great importance when \Ours\ is applied to keypoint detection \cite{tian2019directpose} or instance segmentation \cite{tian2020conditional}.

    \def\cls{ {\rm cls} }
    \def\reg{ {\rm reg} }
    \def\pos{ {\rm pos} }

    \paragraph{Loss Function.} We define our training loss function as follows:
    \begin{align}
    \label{loss_function}
    L(\{\vp_{x, y}\}, \{\vt_{x, y}\}) &= \frac{1}{N_{\pos}} \sum_{x, y}{L_{\cls}(\vp_{x, y}, c^*_{x, y})} \nonumber \\
    & + \frac{\lambda}{N_{\pos}}\sum_{{x, y}}{\mathbbm{1}_{\{c^*_{x, y} > 0\}}L_{\reg}(\vt_{x, y}, \vt^*_{x, y})},
    \end{align}
    where $L_{\cls}$ is focal loss as in \cite{lin2017focal} and $L_{\reg}$ is the GIoU loss \cite{rezatofighi2019generalized}. As shown in experiments, the GIoU loss has better performance than the IoU loss in UnitBox \cite{yu2016unitbox}, which is used in our preliminary version \cite{tian2019fcos}. $N_{\pos}$ denotes the number of positive samples and $\lambda$ being $1$ in this paper is the balance weight for $L_{\reg}$. The summation is calculated  over all locations on the feature maps $F_i$. $\mathbbm{1}_{\{c^*_i > 0\}}$ is the indicator function, being $1$ if $c^*_i > 0$ and $0$ otherwise.

    \paragraph{Inference.} The inference of \Ours\ is straightforward. Given an input images, we forward it through the network and obtain the classification
    scores
    $\vp_{x, y}$
    and the regression prediction $\vt_{x, y}$ for each location on the feature maps $F_i$. Following \cite{lin2017focal}, we choose the location with $p_{x, y} > 0.05$ as positive samples and invert Eq.~(\ref{regression_targets}) to obtain the predicted bounding boxes.

    \subsection{Multi-level Prediction with FPN for \Ours}
    Here we show that how two possible issues of the proposed \Ours\ can be resolved with multi-level prediction with FPN \cite{lin2017feature}.

    First, the large stride (\eg, 16$\times$) of the final feature maps in a CNN can result in a relatively low {\it best possible recall (BPR)\footnote{Upper bound of the recall rate that a detector can achieve.}}. For anchor based detectors, low recall rates due to the large stride can be compensated to some extent by lowering  the IOU score requirements for positive anchor boxes. For \Ours, at the first glance one may think that the BPR can be much lower than anchor-based detectors because it is impossible to recall an object which no location on the final feature maps encodes due to a large stride.
    Here,
    we empirically show that even with a large stride, \Ours\ is still able to produce a good BPR, and it  can even better than the BPR of the anchor-based detector RetinaNet \cite{lin2017focal} in the official implementation Detectron \cite{Detectron2018}
    (refer to Table \ref{table:recall}).
    Therefore, the BPR is actually not a problem of \Ours. Moreover, with multi-level  FPN prediction  \cite{lin2017feature}, the BPR can be improved further to match the best BPR the anchor-based RetinaNet can achieve.

    Second, as shown in Fig.~\ref{fig:training_targets} (right), overlaps in ground-truth boxes can cause
    intractable ambiguity, {\it i.e.}, which bounding box should a location in the overlap regress?
    This ambiguity results in degraded performance. In this work, we show that the ambiguity can be greatly resolved with multi-level prediction (and center sampling), and \Ours\ can obtain {\it on par}, sometimes even better, performance compared with anchor-based ones.

    Specifically, following FPN \cite{lin2017feature}, we detect different size objects on different feature map levels. we make use of five levels of feature maps defined as $\{P_3, P_4, P_5, P_6, P_7\}$. As shown in Fig.~\ref{fig:main_figure}, $P_3$, $P_4$ and $P_5$ are produced by the backbone CNNs' feature maps $C_3$, $C_4$ and $C_5$ with the top-down connections as in \cite{lin2017feature}. $P_6$ and $P_7$ are produced by applying one $3\times3$ convolutional layer with the stride being 2 on $P_5$ and $P_6$, respectively. Note that this is different from the original RetinaNet, which obtain $P_6$ and $P_7$ from the backbone feature maps $C_5$. We find both schemes achieve similar performance but the one we use has fewer parameters. Moreover, the feature levels $P_3$, $P_4$, $P_5$, $P_6$ and $P_7$ have strides 8, 16, 32, 64 and 128, respectively.

    Anchor-based detectors assign different scale anchor boxes to different feature levels. Since anchor boxes and ground-boxes are associated by their IoU scores, this enables different FPN feature levels to handle different scale objects. However, \textit{this couples the sizes of anchor boxes and the target object sizes of each FPN level}, which is problematic. The anchor box sizes should be data-specific, which might be changed from one dataset to another. The target object sizes of each FPN level should depend on the receptive field of the FPN level, which depends on the network architecture. \Ours removes the coupling as we only need focus on the target object sizes of each FPN level and need not design the anchor box sizes. Unlike anchor-based detectors, in \Ours, we directly limit the range of bounding box regression for each level. More specifically, we first compute the regression targets $l^*$, $t^*$, $r ^ *$ and $b^*$ for each location on all feature levels. Next, if a location at feature level $i$ satisfies $\max(l^*$, $ t^*, r^*, b^*) \leq m_{i-1}$ or $\max(l^*, t^*, r^*, b^*) \geq m_i$, it is set as a negative sample and thus not required to regress a bounding box anymore. Here $m_i$ is the maximum distance that feature level $i$ needs to regress. In this work, $m_2$, $m_3$, $m_4$, $m_5$, $m_6$ and $m_7$ are set as $0$, $64$, $128$, $256$, $512$ and $\infty$, respectively. We argue that bounding the maximum distance is a better way to determine the range of target objects for each feature level because this makes sure that the complete objects are always in the receptive field of each feature level. Moreover, since objects of different sizes are assigned to different feature levels and overlapping mostly happens between objects with considerably different sizes, the aforementioned ambiguity can be largely alleviated. If a location, even with multi-level prediction used, is still assigned to more than one ground-truth boxes, we simply choose the ground-truth box with minimal area as its target. As shown in our experiments, with the multi-level prediction, both anchor-free and anchor-based detectors can achieve the same level performance.

    Finally, following \cite{lin2017feature, lin2017focal}, we share the heads between different feature levels,  not only making  the detector parameter-efficient but also improving the detection performance.
    However, we observe that different feature levels are required to regress different size range (\eg, the size range is $[0, 64]$ for $P_3$ and $[64, 128]$ for $P_4$), and therefore it
    may not be the optimal design
    to make use of identical heads for different feature levels. In our %
    preliminary
    version \cite{tian2019fcos}, this issue is %
    addressed
    by multiplying a learnable scalar to the convolutional layer's outputs. In this version, since the regression targets are scaled down by the stride of FPN feature levels, as shown in Eq.~\eqref{regression_targets}, the scalars become less important. However, we still keep them for compatibility.

    \subsection{Center-ness for \Ours}
    After using multi-level prediction, \Ours can already achieve better performance than its anchor-based counterpart RetinaNet. Furthermore, we observed that there are a lot of low-quality detections produced by the locations far away from the center of an object.

\begin{figure}[t]
            \centering
            \includegraphics[width=.2004\textwidth,trim={0 0 0 0},clip]{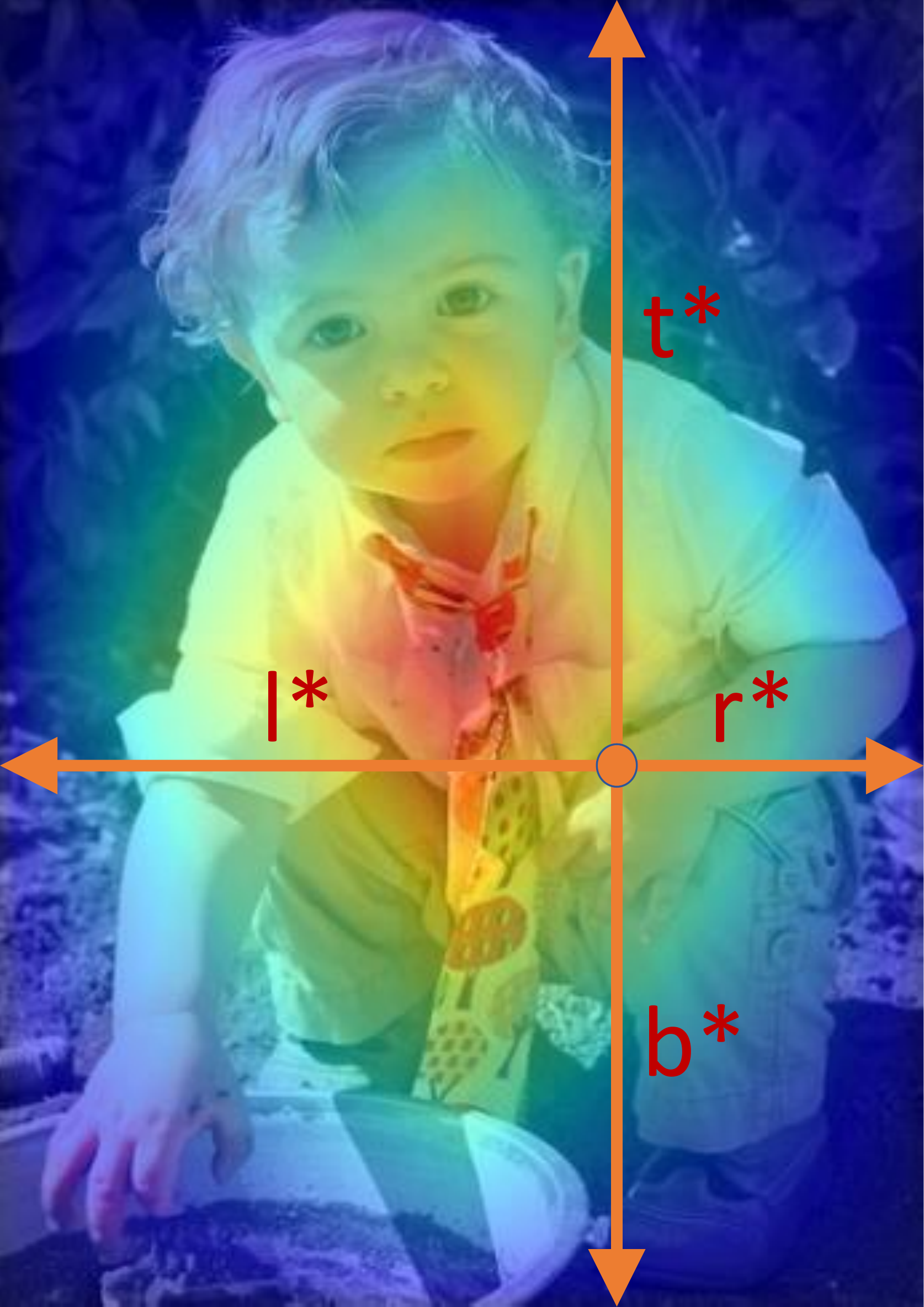}
            \caption{
            \textbf{Center-ness.}
                Red, blue, and other colors denote 1, 0 and the values between them, respectively. Center-ness is computed using  Eq.~(\ref{eq:centerness}) and decays from 1 to 0 as the location deviates from the center of the object. During
                testing,  the center-ness predicted by the network is multiplied with the classification score for NMS, thus
                being able to
                down-weight the low-quality bounding boxes predicted by a location far from the center of an object.}
            \label{fig:centerness}
    \end{figure}

    We propose a simple yet effective strategy to suppress these low-quality detections. Specifically, we add {\it a single-layer branch}, in parallel with the regression branch (as shown in Fig.~\ref{fig:main_figure}) to predict the ``center-ness" of a location\footnote{This is different from our conference version which positions the center-ness on the classification branch, but it has been shown that positioning it on the regression branch can obtain better performance.}. The center-ness depicts the normalized distance from the location to the center of the object that the location is responsible for, as shown Fig.~\ref{fig:centerness}. Given the regression targets $l^*$, $t^*$, $r^*$ and $b^*$ for a location, the center-ness target is defined as,

    \begin{equation}
    \label{eq:centerness}
    \centerness^* = \sqrt{\frac{ \min(l^*, r^*)}{ \max(l^*, r^*)} \times \frac{ \min(t^*, b^*)}{ \max(t^*, b^*)}}.
    \end{equation}
    We employ $\rm sqrt$ here to slow down the decay of the center-ness. The center-ness ranges from $0$ to $1$ and is thus trained with binary cross entropy (BCE) loss. The loss is added to the loss function Eq.~(\ref{loss_function}). When testing, the final score $\vs_{x, y}$ (used for ranking the detections in NMS) is the square root of the product of the predicted center-ness $o_{x, y}$ and the corresponding classification score $\vp_{x, y}$. Formally,
    \begin{equation}
    \label{eq:final_score}
    \vs_{x, y} = \sqrt{\vp_{x, y} \times o_{x, y}},
    \end{equation}
    where $\rm sqrt$ is used to calibrate the order of magnitude of the final score and has no effect on average precision (AP).

    Consequently, center-ness can down-weight the scores of bounding boxes far from the center of an object. As a result, with high probability, these low-quality bounding boxes might be filtered out by the final non-maximum suppression (NMS) process, improving the detection performance \emph{remarkably}.

    \section{Experiments}
    Our experiments are conducted on the large-scale detection benchmark COCO \cite{lin2014microsoft}. Following the common practice \cite{lin2017focal, lin2017feature, ren2015faster}, we use the COCO  \texttt{train2017} split (115K images) for training and \texttt{val2017} split (5K images) as validation for our ablation study. We report our main results on the \texttt{test}-{dev} split (20K images) by uploading our detection results to the evaluation server.

    \paragraph{Training Details.} Unless specified, we use the following implementation details. ResNet-50 \cite{he2016deep} is used as our backbone networks and the same hyper-parameters with RetinaNet \cite{lin2017focal} are used. Specifically, our network is trained with stochastic gradient descent (SGD) for 90k iterations with the initial learning rate being 0.01 and a mini-batch of 16 images. The learning rate is reduced by a factor of 10 at iteration 60k and 80k, respectively. Weight decay and momentum are set as 0.0001 and 0.9, respectively. We initialize our backbone networks with the weights pre-trained on ImageNet \cite{deng2009imagenet}. For the newly added layers, we initialize them as in \cite{lin2017focal}. Unless specified, the input images are resized to have their shorter side being 800 and their longer side less or equal to 1333.

    \paragraph{Inference Details.} We firstly forward the input image through the network and obtain the predicted bounding boxes with the predicted class scores. The next post-processing of \Ours\ exactly follows that of RetinaNet \cite{lin2017focal}. The post-processing hyper-parameters are also the same except that we use NMS threshold $0.6$ instead of $0.5$ in RetinaNet. Experiments will be conducted to show the effect of the NMS threshold. Moreover, we use the same sizes of input images as in training.

    \subsection{Analysis of \Ours}
    \subsubsection{Best Possible Recall (BPR) of \Ours}
    \begin{table}[!t]
        \centering
        \small
        \begin{tabular}{ l|c|c|c }
            Method & w/ FPN & Low-quality matches & BPR (\%) \\
            \Xhline{2\arrayrulewidth}
            RetinaNet & \checkmark & Not used & 88.16 \\
            RetinaNet & \checkmark & $\geq 0.4$ & 91.94 \\
            RetinaNet & \checkmark & All & \textbf{99.32} \\
            \hline
            \Ours &  & - & 96.34 \\
            \Ours & \checkmark & - & 98.95 \\
        \end{tabular}
        \caption{The best possible recall (BPR) of anchor-based RetinaNet under a variety of matching rules and the BPR of \Ours on the COCO \texttt{val2017} split. \Ours\ has very similar BPR to the best anchor-based one and has much higher recall than the official implementation in \texttt{Detectron} \cite{Detectron2018}, where only low-quality matches with IOU $\geq0.4$ are considered.}
        \label{table:recall}
    \end{table}

    \begin{table}[!b]
        \centering
        \small
        \begin{tabular}{ c|c|c|c|c}
            w/ ctr. sampling & w/ FPN & $1$ & $2$ & $\geq 3$ \\
            \Xhline{2\arrayrulewidth}
             & & $76.60\%$ & $20.05\%$ & $3.35\%$ \\
            & \checkmark & $92.58\%$ & $6.97\%$ & $0.45\%$ \\
            \hline
            \checkmark & & $96.52\%$ & $3.34\%$ & $0.14\%$ \\
            \checkmark & \checkmark & $97.34\%$ & $2.59\%$ & $0.07\%$ \\
        \end{tabular}
        \caption{
            The ratios of the ambiguous samples to all the positive samples in \Ours. $1, 2$ and $\geq 3$ denote the number of ground-truth boxes a location should be associated to. If the number is greater than $1$, the location is defined as an ``ambiguous sample" in this work. As shown in the table, with center sampling and FPN, the ratio of ambiguous samples is low (\ie, $<3\%$).
        }
        \label{table:overlapping}
    \end{table}

    \begin{table*}[!tbh]
        \centering
        \small
        \begin{tabular}{ l | c | c c | c c c | c c c | c c c}
            Method & AP & AP$_{50}$ & AP$_{75}$ & AP$_{S}$ & AP$_{M}$ & AP$_{L}$ & AR$_{1}$ & AR$_{10}$ & AR$_{100}$ & AR$_{S}$ & AR$_{M}$ & AR$_{L}$ \\
            \Xhline{2\arrayrulewidth}
            RetinaNet (\#A=9) & 35.9 & 55.8 & 38.4 & 20.6 & 39.8 & 46.6 & 31.0 & 49.8 & 53.0 & 33.8 & 57.4 & 67.9 \\ %
            \hline
            RetinaNet (\#A=1) w/ imprv. & 35.2 & 55.6 & 37.0 & 19.9 & 39.2 & 45.2 & 30.4 & 49.9 & 53.5 & 33.6 & 57.7 & 68.2 \\ %
            RetinaNet (\#A=9) w/ imprv. & 37.6 & 56.6 & 40.6 & 21.5 & 42.1 & 48.0 & 32.1 & 52.2 & 56.4 & 35.5 & 60.2 & 72.7 \\ %
            \hline
            \Ours\ w/o ctr.-ness &  38.0 & 57.2 & 40.9 & 21.5 & 42.4 & 49.1 & 32.1 & 52.4 & 56.2 & 36.6 & 60.6 & 71.9  \\
            \Ours\ w/ ctr.-ness & \textbf{38.9} & \textit{57.5} & \textbf{42.2} & \textbf{23.1} & \textbf{42.7} & \textbf{50.2} & \textbf{32.4} & \textbf{53.8} & \textbf{57.5} & \textbf{38.5} & \textbf{62.1} & \textbf{72.9} \\
        \end{tabular}
        \caption{\textbf{\Ours\ vs.\ RetinaNet } on \texttt{val2017} split with ResNet-50-FPN as the backbone. All experiments use the same training settings. The proposed  anchor-free \Ours\ achieves even better performance than anchor-based RetinaNet. \#A: the number of anchors per location. RetinaNet (\#A=9): the original RetinaNet from \texttt{Detectron2} \cite{wu2019detectron2}. RetinaNet w/ imprv. RetinaNet with the universal improvements in \Ours including Group Normalization (GN) \cite{wu2018group}, GIoU loss \cite{fcosplus} and scalars in regression, using $P_5$ instead of $C_5$ and NMS threshold $0.6$ instead of $0.5$. We have tried our best to make all the details consistent. As shown the table, even without the center-ness branch, the much simpler \Ours\ already outperforms "RetinaNet (\#A=9) w/ imprv" by $0.4\%$ in AP. With the center-ness branch, the performance is further improved to $38.9\%$ in AP.}
        \label{table:ours_vs_anchor_based_detector}
    \end{table*}

    We first address the concern that is \Ours might not provide a good best possible recall (BPR) (\ie, upper bound of the recall rate). In the section, we show that the concern is not necessary by comparing BPR of \Ours and that of its anchor-based counterpart RetinaNet on the COCO \texttt{val2017} split. The following analyses are based on the \Ours implementation in \texttt{AdelaiDet}. Formally, BPR is defined as the ratio of the number of ground-truth boxes that a detector can recall at the most to the number of all ground-truth boxes. A ground-truth box is considered recalled if the box is assigned to at least one training sample (\ie, a location in \Ours\ or an anchor box in anchor-based detectors), and a training sampling can be associated to \emph{at most} one ground-truth box. As shown in Table~\ref{table:recall}, both with FPN, \Ours and RetinaNet obtain similar BPR ($98.95$ vs $99.32$)\footnote{One might think
    that
    the BPR of RetinaNet should be $1$ if all the low-quality matches are used. However, this is not true in some cases as each anchor can only be associated to the ground-truth box with the highest IOU to it. For example, if two boxes A and B, both of which are small and contained in the common of all the anchor boxes at the same location.
    Clearly,
    for all these anchor boxes, the box with larger area has higher IOU scores and thus all the anchor boxes will be associated to it. Another one will be missing.}. Due to the fact that the best recall of current detectors are much lower than $90\%$, the small BPR gap (less than $0.5\%$) between \Ours\ and the anchor-based RetinaNet will not actually affect the performance of a detector. It is also confirmed in Table~\ref{table:ours_vs_anchor_based_detector}, where \Ours\ achieves better or similar AR than RetinaNet under the same training and testing settings. Even more surprisingly, only with feature level $P_4$ with stride being 16 (\ie, no FPN), \Ours\ can obtain a decent BPR of $96.34\%$. The BPR is much higher than the BPR of $91.94\%$ of the RetinaNet in the official implementation \texttt{Detectron} \cite{Detectron2018}, where only the low-quality matches with IOU $\geq0.4$ are used. Therefore, the concern about low BPR may not be necessary.

    \subsubsection{Ambiguous Samples in \Ours}
    Another concern about the FCN-based detector is that it may have a large number of \emph{ambiguous samples} due to the overlap in ground-truth boxes, as shown in Fig.~\ref{fig:training_targets} (right). In Table~\ref{table:overlapping}, we show the ratios of the ambiguous samples to all positive samples on \texttt{val2017} split. If a location should be associated to multiple ground-truth boxes without using the rule of choosing the box with minimum area, the location is defined as an ``ambiguous sample". As shown in the table, there are indeed a large amount of ambiguous samples ($23.40\%$) if FPN is not used (\ie, only $P_4$ used). However, with FPN, the ratio can be significantly reduced to only $7.42\%$ since most of overlapped objects are assigned to different feature levels. Furthermore, if the center sampling is used, the ambiguous samples can be significantly reduced. As shown in Table~\ref{table:overlapping}, even without FPN, th ratio is only $3.48\%$. By further applying FPN, the ratio is reduced to $2.66\%$. Note that it does not imply that there are $2.66\%$ locations where \Ours\ makes mistakes. As mentioned before, these locations are associated with the smallest one among the ground-truth boxes associated to the same location. Therefore, these locations only take the risk of missing some larger objects. In other words, it may harm the recall of \Ours. However, as shown in Table~\ref{table:recall}, the recall gap between \Ours\ and RetinaNet is negligible, which suggests that the ratio of the missing objects is extremely low.

    \subsubsection{The Effect of Center-ness}
    \begin{figure}[t!]
        \centering
        \includegraphics[width=\linewidth]{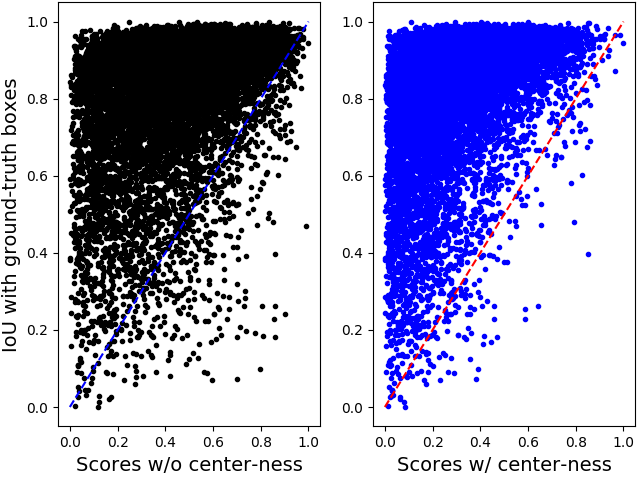}
        \caption{\textbf{Quantitative results of applying the center-ness scores to classification scores}. A point in the figure denotes a bounding box. The dashed line is the line $y = x$. As shown in the right figure, after applying the center-ness scores, the boxes with low IoU scores but high confidence scores (\ie, under the line $ y = x $) are reduced substantially.}
        \label{fig:vis_ctrness}
    \end{figure}

    \begin{table}[b]
        \setlength{\tabcolsep}{4.5pt}
        \centering
        \small
        \begin{tabular}{ l|c|c c|c c c }
            & AP & AP$_{50}$ & AP$_{75}$ & AP$_{S}$ & AP$_{M}$ & AP$_{L}$ \\
            \Xhline{2\arrayrulewidth}
            w/o ctr.-ness & 38.0 & 57.2 & 40.9 & 21.5 & 42.4 & 49.1 \\
            w/ ctr.-ness$^\dag$ & 37.5 & 56.5 & 40.2 & 21.6 & 41.5 & 48.5 \\
            w/ ctr.-ness (L1) & \textbf{38.9} & \textbf{57.6} & 42.0 & 23.0 & 42.3 & \textbf{51.0} \\
            w/ ctr.-ness & \textbf{38.9} & 57.5 & \textbf{42.2} & \textbf{23.1} & \textbf{42.7} & \textbf{50.2} \\
        \end{tabular}
        \caption{\textbf{Ablation study for the proposed center-ness branch} on the \texttt{val2017} split. ctr.-ness$^\dag$: using the center-ness computed from the predicted regression vector when testing (\ie, replacing the ground-truth values with the predicted ones in Eq.~(\ref{eq:centerness})). ``ctr.-ness" is that using center-ness predicted from the proposed center-ness branch. The center-ness branch improves the detection performance. On the contrary, using ``ctr.-ness$^\dag$" even degrades the performance, which suggests that the separate center-ness branch is necessary. w/ ctr.-ness (L1): using L1 instead of BCE as the loss to optimize the center-ness.}
        \label{table:center-ness}
    \end{table}
    As mentioned before, we propose ``center-ness" to suppress the low-quality detected bounding boxes produced by the locations far from the center of an object. As shown in Table~\ref{table:center-ness}, the center-ness branch can boost AP from $38.0\%$ to $38.9\%$. Compared to our conference version \cite{tian2019fcos}, the gap is relatively smaller since we make use of the center sampling by default and it already eliminates a large number of false positives. However, the improvement is still impressive as the center-ness branch only adds negligible computational time. Moreover, we will show later that the center-ness can bring a large improvement in crowded scenarios. One may note that center-ness can also be computed with the predicted regression vector without introducing the extra center-ness branch. However, as shown in Table~\ref{table:center-ness}, the center-ness computed from the regression vector cannot improve the performance and thus the separate center-ness is necessary.

    We visualize the effect of applying the center-ness in Fig.~\ref{fig:centerness}. As shown in the figure, after applying the center-ness scores to the classification scores, the boxes with low IoU scores but high confidence scores are largely eliminated (\ie, the points under the line $y = x$ in the Fig.~\ref{fig:centerness}), which are potential false positives.

        \begin{table}[t!]
        \centering
        \small
        \begin{tabular}{l | c | c c | c c c}
            & AP & AP$_{50}$ & AP$_{75}$ & AP$_{S}$ & AP$_{M}$ & AP$_{L}$ \\
            \Xhline{2\arrayrulewidth}
            Baseline & \textbf{38.9} & \textbf{57.5} & \textbf{42.2} & \textbf{23.1} & \textbf{42.7} & \textbf{50.2} \\
            \hline
            w/o GN & 37.9 & 56.4 & 40.9 & 22.1 & 41.8 & 48.8 \\
            w/ IoU & 38.6 & 57.2 & 41.9 & 22.4 & 42.1 & 49.8 \\
            w/ C$_5$ & 38.5 & 57.4 & 41.7 & 22.8 & 42.1 & 49.3 \\
        \end{tabular}
        \caption{Ablation study for design choices in \Ours. w/o GN:
            without using
            Group Normalization (GN) for the convolutional layers in heads.
            w/ IoU: using IoU loss in \cite{yu2016unitbox} instead of GIoU. w/ $C_5$: using $C_5$ instead of $P_5$.
        }
        \label{table:other_design_choices}
    \end{table}

    \begin{table}[t!]
        \setlength{\tabcolsep}{7pt}
        \centering
        \small
        \begin{tabular}{c | c | c c | c c c}
            $r$ & AP & AP$_{50}$ & AP$_{75}$ & AP$_{S}$ & AP$_{M}$ & AP$_{L}$ \\
            \Xhline{2\arrayrulewidth}
            $1.0$ & 38.5 & 57.2 & 41.5 & 22.6 & 42.3 & 49.7 \\
            $1.5$ & \textbf{38.9} & 57.5 & \textbf{42.2} & \textbf{23.1} & \textbf{42.7} & \textbf{50.2} \\
            $2.0$ & 38.8 & \textbf{57.7} & 41.7 & 22.7 & 42.6 & 49.9 \\
        \end{tabular}
        \caption{Ablation study for the radius $r$ of positive sample regions (defined in Section~\ref{sec:main_method}).
        }
        \label{table:radius}
    \end{table}

    \begin{table}
        \setlength{\tabcolsep}{4.5pt}
        \centering
        \small
        \begin{tabular}{ l|c|c c|c c c }
            Strategy & AP & AP$_{50}$ & AP$_{75}$ & AP$_{S}$ & AP$_{M}$ & AP$_{L}$ \\
            \Xhline{2\arrayrulewidth}
            FPN & 37.7 & 56.6 & 40.6 & 22.2 & 40.9 & 49.7 \\
            $\sqrt{(h^* \times w^*)} / 2$ & 37.6 & 56.5 & 40.6 & 22.4 & 41.6 & 47.3 \\
            ${\rm max}(h^*, w^*) / 2$ & 38.1 & 57.0 & 41.3 & 22.5 & 41.8 & 48.7 \\
            ${\rm max}(l^*, t^*, r^*, b^*)$ & \textbf{38.9} & \textbf{57.5} & \textbf{42.2} & \textbf{23.1} & \textbf{42.7} & \textbf{50.2} \\
        \end{tabular}
        \caption{Ablation study for different strategies of assigning objects to FPN levels. FPN: the strategy of assigning object proposals (\ie, ROIs) to FPN levels in the original FPN, described in the text. $h^*$ and $w^*$ are the height and width of a ground-truth box, respectively. $l^*$, $t^*$, $r^*$ and $b^*$ are the distances from a location to the four boundaries of a ground-truth box. ``$\rm max$$(l^*, t^*, r^*, b^*)$" (used by \Ours) has the best performance.}
        \label{table:assignment}
    \end{table}

    \subsubsection{Other Design Choices}

    Other design choices are also investigated. As shown Table~\ref{table:other_design_choices}, removing group normalization (GN) \cite{wu2018group} in both the classification and regression heads drops the performance by $1\%$ AP. By replacing GIoU \cite{rezatofighi2019generalized} with the origin IoU loss in \cite{yu2016unitbox}, the performance drops by $0.3\%$ AP. Using $C_5$ instead of $P_5$ also degrades the performance. Moreover, using $P_5$ can reduce the number of the network parameters. We also conduct experiments for the radius $r$ of positive sample regions. As shown in Table~\ref{table:radius}, $r = 1.5$ has the best performance on COCO \texttt{val} split. We also attempted to change the sampling area from square sub-boxes to the rectangle sub-boxes with the same aspect ratio as the ground-truth boxes, which results in similar performance. It suggests that the shape of the sampling area may be not sensitive to the final performance.

    We also conduct experiments with different strategies of assigning objects to FPN levels. First, we experiment with the assigning strategy when FPN \cite{lin2017feature} assigns the object proposals (\ie, ROIs) to FPN levels. It assigns the objects according to the formulation $k = \floor{k_0 + {\rm log}_2({\sqrt{wh} / 224})}$, where $k \in \{3, 4, 5, 6, 7\}$ is the target FPN level, $w$ and $h$ are the ground-truth box's width and height, respectively, and $k_0$ is the target level which an object with scale $224$ should be mapped into. We use $k_0 = 5$. As shown in Table~\ref{table:assignment}, this strategy results in degraded performance ($37.7\%$ AP). We conjecture that it may be because the strategy cannot make sure the complete object
    be
    within the receptive field of the target FPN level.

    Similarly, $\sqrt{(h^* \times w^*)} / 2$ and ${\rm max}(h^*, w^*) / 2$ also
    deteriorate
    the performance. Eventually, ${\rm max}(l^*, t^*, r^*, b^*)$ achieves the best performance as the strategy makes sure that the complete target objects are always in the effective receptive field of the FPN level. Moreover, this implies that the range hyper-parameters of each FPN level (\ie, $m_i$) is mainly related to the network architecture (which determines the receptive fields). This is a desirable feature since it eliminates the hyper-parameter tuning when \Ours is applied to different datasets.

    \subsection{\Ours\ vs.\ Anchor-based Counterparts}
    Here, we compare \Ours\ with its anchor-based counterpart RetinaNet on the challenging benchmark COCO, demonstrating that the much simpler anchor-free \Ours\ is superior.

    In order to make a fair comparison, we add the universal improvements in \Ours to RetinaNet. The improved RetinaNet is denoted as ``\textit{RetinaNet w/ imprv.}" in Table~\ref{table:ours_vs_anchor_based_detector}. As shown the table, even without the center-ness branch, \Ours\ achieves $0.4\%$ better AP than ``RetinaNet (\#A=9) w/ imprv." ($38.0\%$ vs $37.6\%$ in AP). The performance of \Ours\ can be further boosted to $38.9\%$ with the help of the proposed center-ness branch. Moreover, it is worth noting that \Ours\ achieves much better performance than the RetinaNet with a single anchor per location ``RetinaNet (\#A=1) w/ imprv." ($38.0\%$ vs $35.2\%$), which suggests that \Ours\ is not equivalent to the single-anchor RetinaNet. The major difference is \Ours\ does not employ IoU scores between anchor boxes and ground-truth boxes to determine the training labels.

    Given the superior performance and merits of the anchor-free detector (\eg, much simpler and fewer hyper-parameters), we encourage the community to rethink the necessity of anchor boxes in object detection.
    \subsection{Comparison with State-of-the-art Detectors on COCO}\label{sec:state_of_the_art}
        \begin{table*}[!htb]
        \centering
        \small
        \begin{tabular}{ l | l |c c c | c c c }
            Method & Backbone & AP & AP$_{50}$ & AP$_{75}$ & AP$_{S}$ & AP$_{M}$ & AP$_{L}$ \\
            \Xhline{2\arrayrulewidth}
            \hspace{-0.15cm}\textrm{Two-stage methods:} &&&&&&& \\
            Faster R-CNN+++ \cite{he2016deep} & ResNet-101 & 34.9 & 55.7 & 37.4 & 15.6 & 38.7 & 50.9 \\
            Faster R-CNN w/ FPN \cite{lin2017feature} & ResNet-101-FPN & 36.2 & 59.1 & 39.0 & 18.2 & 39.0 & 48.2 \\
            Faster R-CNN by G-RMI \cite{huang2017speed} & Inception-ResNet-v2 \cite{szegedy2017inception} & 34.7 & 55.5 & 36.7 & 13.5 & 38.1 & 52.0 \\
            Faster R-CNN w/ TDM \cite{shrivastava2016beyond} & Inception-ResNet-v2-TDM & 36.8 & 57.7 & 39.2 & 16.2 & 39.8 & 52.1 \\
            \hline
            \hspace{-0.15cm}\textrm{One-stage methods:} &&&&&&& \\
            YOLOv2 \cite{redmon2017yolo9000} & DarkNet-19 \cite{redmon2017yolo9000} & 21.6 & 44.0 & 19.2 & 5.0 & 22.4 & 35.5 \\
            SSD513 \cite{liu2016ssd} & ResNet-101-SSD & 31.2 & 50.4 & 33.3 & 10.2 & 34.5 & 49.8 \\
            YOLOv3 $608\times608$ \cite{redmon2018yolov3} & Darknet-53 & 33.0 & 57.9 & 34.4 & 18.3 & 35.4 & 41.9 \\
            DSSD513 \cite{fu2017dssd} & ResNet-101-DSSD & 33.2 & 53.3 & 35.2 & 13.0 & 35.4 & 51.1 \\
            RetinaNet \cite{lin2017focal} & ResNet-101-FPN & 39.1 & 59.1 & 42.3 & 21.8 & 42.7 & 50.2 \\
            CornerNet \cite{law2018cornernet} & Hourglass-104 & 40.5 & 56.5 & 43.1 & 19.4 & 42.7 & 53.9 \\
            FSAF \cite{zhu2019feature} & ResNeXt-64x4d-101-FPN & 42.9 & 63.8 & 46.3 & 26.6 & 46.2 & 52.7 \\
            CenterNet511 \cite{duan2019centernet} & Hourglass-104 & 44.9 & 62.4 & 48.1 & 25.6 & 47.4 & 57.4 \\
            \hline
            \Ours & ResNet-101-FPN & 43.2 & 62.4 & 46.8 & 26.1 & 46.2 & 52.8 \\
            \Ours & ResNeXt-32x8d-101-FPN & 44.1 & 63.7 & 47.9 & 27.4 & 46.8 & 53.7 \\
            \Ours & ResNeXt-64x4d-101-FPN & 44.8 & 64.4 & 48.5 & 27.7 & 47.4 & 55.0 \\
            \Ours w/ deform.\  conv.\  v2 \cite{zhu2019deformable} & ResNeXt-32x8d-101-FPN & 46.6 & 65.9 & 50.8 & 28.6 & 49.1 & 58.6 \\
            \hline
            \Ours & ResNet-101-BiFPN \cite{tan2019efficientdet} & 45.0 & 63.6 & 48.7 & 27.0 & 47.9 & 55.9 \\
            \Ours & ResNeXt-32x8d-101-BiFPN & 46.2 & 65.2 & 50.0 & 28.7 & 49.1 & 56.5 \\
            \Ours w/ deform.\  conv.\  v2 & ResNeXt-32x8d-101-BiFPN & 47.9 & 66.9 & 51.9 & 30.2 & 50.3 & 59.9 \\
            \hline
            \hspace{-0.215cm}
            \textrm{w/ test-time augmentation:} &&&&&&& \\
            \Ours & ResNet-101-FPN & 45.9 & 64.5 & 50.4 & 29.4 & 48.3 & 56.1 \\
            \Ours & ResNeXt-32x8d-101-FPN & 47.0 & 66.0 & 51.6 & 30.7 & 49.4 & 57.1\\
            \Ours & ResNeXt-64x4d-101-FPN & 47.5 & 66.4 & 51.9 & 31.4 & 49.7 & 58.2 \\
            \Ours w/ deform.\  conv.\  v2 & ResNeXt-32x8d-101-FPN & 49.1 & 68.0 & 53.9 & 31.7 & 51.6 & 61.0 \\
            \hline
            \Ours & ResNet-101-BiFPN & 47.9 & 65.9 & 52.5 & 31.0 & 50.7 & 59.7 \\
            \Ours & ResNeXt-32x8d-101-BiFPN & 49.0 & 67.4 & 53.6 & 32.0 & 51.7 & 60.5 \\
            \Ours w/ deform.\  conv.\  v2 & ResNeXt-32x8d-101-BiFPN & \textbf{50.4} & \textbf{68.9} & \textbf{55.0} & \textbf{33.2} & \textbf{53.0} & \textbf{62.7} \\

        \end{tabular}
        \caption{\Ours\ vs.\ other state-of-the-art two-stage or one-stage detectors (\emph{single-model results}). \Ours\ outperforms a few recent
        anchor-based
        and
        anchor-free detectors by a
        considerable
        margin.
        }
        \label{table:ours_vs_stoa_detectors}
    \end{table*}
    We compare \Ours\ with other state-of-the-art object detectors on \texttt{test}-\texttt{dev} split of MS-COCO benchmark. For these experiments, following previous works \cite{lin2017focal, liu2016ssd}, we make use of multi-scale training. To be specific, during training, the shorter side of the input image is sampled from $[640, 800]$ with a step of $32$. Moreover, we double the number of iterations to 180K (with the learning rate change points scaled proportionally). Other settings are exactly the same as the model with AP $38.9\%$ on \texttt{val2017} in Table~\ref{table:ours_vs_anchor_based_detector}.

\begin{figure*}[t!]
        \centering
        \includegraphics[width=0.99\linewidth]{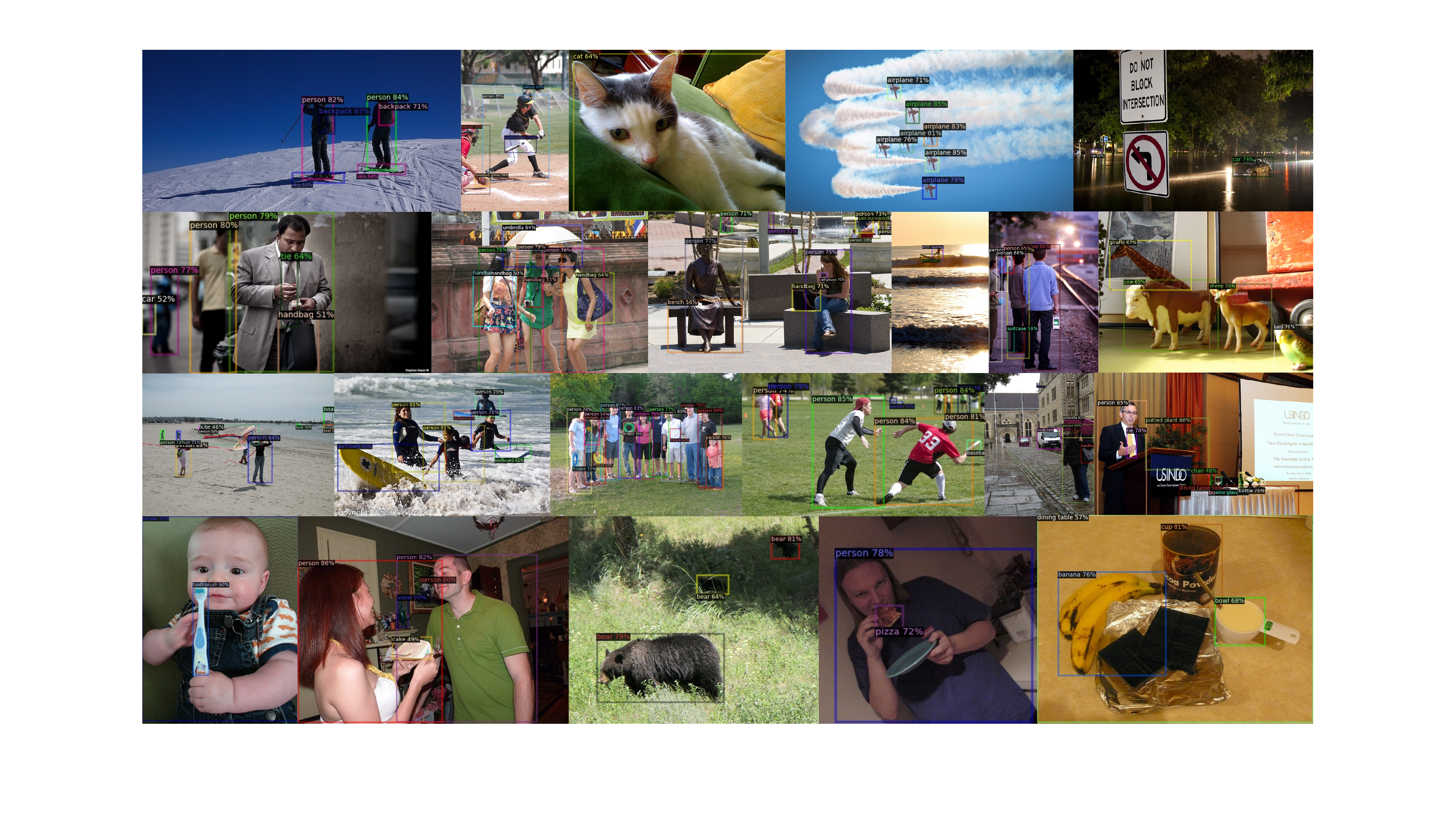}
        \caption{%
        Qualitative
        results. As shown in the figure, \Ours\ works well with
            a wide range of objects including crowded, occluded, extremely small and very large objects.}
        \label{fig:vis}
    \end{figure*}

    As shown in Table \ref{table:ours_vs_stoa_detectors}, with ResNet-101-FPN, \Ours\ outperforms the original RetinaNet with the same backbone by $4.1\%$ AP ($43.2\%$ vs $39.1\%$). Compared to other one-stage detectors such as SSD \cite{liu2016ssd} and DSSD \cite{fu2017dssd}, we also achieve much better performance. Moreover, \Ours\ also surpasses the classical two-stage anchor-based detector Faster R-CNN by a large margin ($43.2\%$ vs.\  $36.2\%$). To our knowledge, it is the first time that an anchor-free detector, without any bells and whistles, outperforms anchor-based detectors by a large margin. Moreover, \Ours\ also outperforms the previous anchor-free detector CornerNet \cite{law2018cornernet} and CenterNet \cite{duan2019centernet} while being much simpler since they requires to group corners with embedding vectors, which needs special design for the detector. Thus, we argue that \Ours\ is more likely to serve as a strong and simple alternative to current mainstream anchor-based detectors. Quantitative results are shown in Fig.~\ref{fig:vis}.
    It appears that
    \Ours works well with a variety of challenging cases.

    We also introduce some complementary techniques to \Ours. First, deformable convolutions are used in stages $3$ and $4$ of the backbone, and also replace the last convolutional layers in the classification and regression towers (\ie, the $4\times$ convolutions shown in Fig.~\ref{fig:main_figure}). As shown in Table~\ref{table:ours_vs_stoa_detectors}, by applying deformable convolutions \cite{dai2017deformable, zhu2019deformable} to ResNeXt-32x8d-101-FPN based \Ours, the performance is improved from $44.1\%$ to $46.6\%$ AP, as shown in Table \ref{table:ours_vs_stoa_detectors}. In addition, we also attempt to replace FPN in \Ours with BiFPN \cite{tan2019efficientdet}. We make use of BiFPN in D3 model in \cite{tan2019efficientdet}. To be specific, the single cell of BiFPN is repeated $6$ times and the number of its output channels is set to $160$. Note that unlike the original BiFPN, we do not employ depthwise separable convolutions in it. As a result, BiFPN generally improves all \Ours\ models by $\sim 2\%$ AP and pushes the performance of the best model to $47.9\%$.

    We also report the result of
    using
    test-time data augmentation. Specifically, in inference, the input image is respectively resized to $[400, 1200]$ pixels with step $100$. At each scale, the original image and its horizontal flip are evaluated. The results from these augmented images are merged by NMS. As shown in Table \ref{table:ours_vs_stoa_detectors}, the test-time augmentation improves the best performance to $50.4\%$ AP.

    \subsection{Real-time \Ours}
    \begin{table}
        \centering
        \small
        \begin{tabular}{ l|c|c|c}
            Method & FPS & AP & AP \texttt{test}-\texttt{dev} \\
            \Xhline{2\arrayrulewidth}
            YOLOv3 (Darknet-53) \cite{redmon2018yolov3} & 26 & $-$  & 33.0 \\
            CenterNet (DLA-34) \cite{zhou2019objects} & \textbf{52} & 37.4 & 37.3 \\
            \hline
            FCOS-RT (R-50) & 38 & 40.2 & 40.2 \\
            FCOS-RT (DLA-34-BiFPN) & 43 & \textbf{42.1} & \textbf{42.2} \\
            FCOS-RT (DLA-34) & 46 & 40.3 & 40.3 \\
            FCOS-RT w/ shtw. (DLA-34) & \textbf{52} & 39.1 & 39.2 \\
        \end{tabular}
        \caption{\textbf{Real-time \Ours (FCOS-RT) models.} AP is on COCO \texttt{val} split. ``shtw.": sharing towers (\ie, $4\times$ conv.\ layers shown in Fig.~\ref{fig:main_figure}) between the classification and regression branches. The inference time is measured with a single 1080Ti  or Titan XP GPU (these two GPUs' speeds are close).
        FPS of all  FCOS models is measured on a 1080Ti GPU.
        We measure the FPS of YOLOv3 on 1080Ti
        using the
        code released by
        \cite{redmon2018yolov3}. For \Ours and YOLOv3, we measure the end-to-end inference time here (\ie, from prepossessing to the final output boxes).
        }
        \label{table:real_time}
    \end{table}
    We also design a real-time version FCOS-RT. In the real-time settings, we reduce the shorter side of input images from $800$ to $512$ and the maximum longer size from $1333$ to $736$, which decreases the inference time per image by $\sim 50\%$. With the smaller input size, the higher feature levels $P_6$ and $P_7$ become less important. Thus, following BlendMask-RT \cite{chen2020blendmask}, we remove $P_6$ and $P_7$, further reducing the inference time. Moreover, in order to boost the performance of the real-time version, we employ a more aggressive training strategy. Specifically, during training, multi-scale data augmentation is used and the shorter size of input image is sampled from $256$ to $608$ with interval $32$. Synchronized batch normalization (SyncBN) is used. We also increase the training iterations to $360K$ (\ie, $4\times$). The learning rate is decreased by a factor of $10$ at iteration $300K$ and $340K$.

    The resulting real-time models are shown in Table~\ref{table:real_time}. With ResNet-50, FCOS-RT can achieve $40.2\%$ AP at $38$ FPS on a single 1080Ti GPU card. We further replace ResNet-50 with the backbone DLA-34 \cite{yu2018deep}, which results in a better speed/accuracy trade-off ($40.3\%$ AP at $46$ FPS). In order to compare with CenterNet\cite{zhou2019objects}, we share the towers (\ie, $4\times$ conv.\ layers shown in Fig.~\ref{fig:main_figure}) between the classification and regression branches, which improves the speed from $46$ FPS to $52$ FPS but
    deteriorate
    the performance by $1.2\%$ AP. However, as shown in Table~\ref{table:real_time}, the model still outperforms CenterNet \cite{zhou2019objects} by $1.7\%$ AP at the same speed.
    For  the real-time models, we also
    replace FPN with BiFPN as in Section~\ref{sec:state_of_the_art}, resulting $1.8\%$ AP improvement (from $40.3\%$ to $42.1\%$) at similar speed. %
    A speed/accuracy comparison between \Ours\ and a few recent detection
    methods is shown in Fig.~\ref{fig:speed_acc}.

    \subsection{\Ours\ on CrowdHuman}
        \begin{figure*}[t!]
        \centering
        \includegraphics[width=0.99\linewidth]{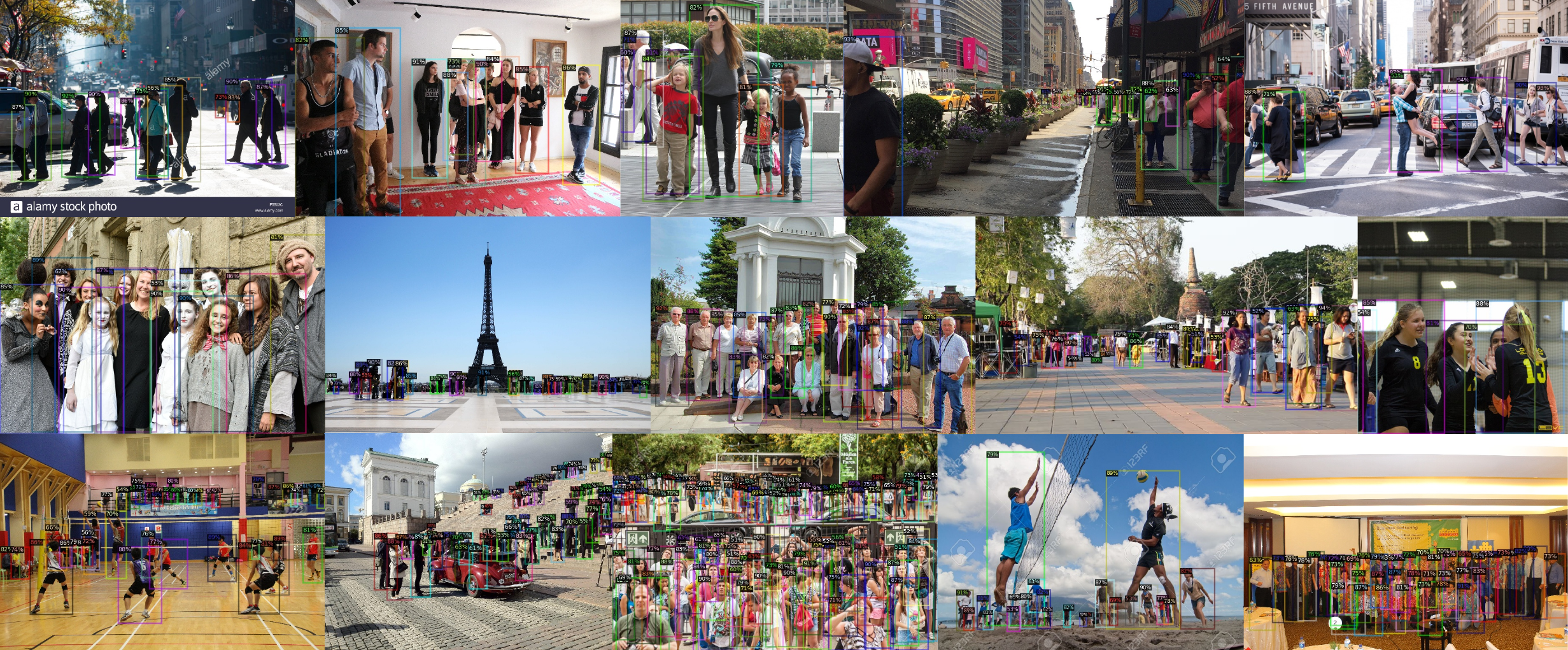}
        \caption{
        Qualitative
        results on the CrowdHuman \texttt{val} set with the ResNet-50-FPN backbone.}
        \label{fig:vis_crowdhuman}
    \end{figure*}

        \begin{table}[b!]
        \setlength{\tabcolsep}{12pt}
        \centering
        \small
        \begin{tabular}{l | c c c}
            & AP & MR$^{-2}$ & JI \\
            \Xhline{2\arrayrulewidth}
            RetinaNet w/ imprv. & 81.60 & 57.36 & 72.88 \\
            \Ours\ w/o ctr.-ness & 83.16 & 59.04 & 73.09 \\
            \Ours\ w/ ctr.-ness & 85.0 & 51.34 & 74.97 \\
            \hline
            ~ ~ ~ $+$ MIP ($K=2$) & 85.19 & 51.60 & 75.14 \\
            ~ ~ ~ $+$ Set NMS  \cite{chu2020detection} & \textbf{87.28} & \textbf{51.21} & \textbf{77.34} \\
        \end{tabular}
        \caption{\textbf{\Ours\ for crowded object detection on the CrowdHuman dataset}. Even on the highly crowded benchmark, \Ours still attains even better performance than anchor-based RetinaNet. Note that lower MR$^{-2}$ is better. ``MIP w.\ set NMS'': Multiple Instance Prediction, which predicts multiple instances from a single location as proposed by \cite{chu2020detection}. Note that we are not pursuing the state-of-the-art performance on the benchmark. We only show that the anchor boxes are not necessary even on the highly-crowded benchmark.}
        \label{table:fcos_crowd_human}
    \end{table}
    We also conduct experiments on the highly crowded dataset CrowdHuman \cite{shao2018crowdhuman}. CrowdHuman consists of $15k$ images for training, $4, 370$ for validation and $5, 000$ images for testing. Following previous works on crowded benchmark \cite{chu2020detection, shao2018crowdhuman}, we use AP, \textit{long-average Miss Rate on False Positive Per Image in $[10^{-2}, 100]$} (MR$^{-2}$) \cite{dollar2011pedestrian} and \textit{Jaccard Index} (JI) as the evaluation metrics. Note that lower MR$^{-2}$ is better. Following \cite{chu2020detection}, all experiments here are trained on the \texttt{train} split for $30$ epochs with batch size $16$ and then evaluated on the \texttt{val} split. Two some changes are made when \Ours is applied to the benchmark. First, the NMS threshold is set as $0.5$ instead of $0.6$. We find that it has large impact on MR$^{-2}$ and JI. Second, when a location is supposed to be associated to multiple ground-truth boxes, on COCO, we choose the object with minimal area as the target for the location. On CrowdHuman, we instead choose the target with minimal distance to the location. The distance between a location and an object is defined as the distance from the location to the center of the object. On COCO, both schemes result in similar performance. However, the latter has much better performance than the former on the highly crowded dataset. Other settings are the same as that of COCO.

    First, we count the ambiguous sample ratios on CrowdHuman \texttt{val} set. With FPN-based \Ours, there are $84.47\%$ unambiguous positive samples (with one ground-truth box), $13.63\%$ with two ground-truth boxes, $1.69\%$ with three ground-truth boxes and the rest ($<0.3\%$) with more than three ground-truth boxes. Given the much higher ambiguous sample ratio than COCO, it is expected that \Ours\ will have inferior performance on the highly crowded dataset.

    We compare \Ours\ without center-ness with the the improved RetinaNet (\ie, ``RetinaNet w/ imprv."). To our surprise, even without center-ness, \Ours\ can already achieve decent performance. As shown in Table~\ref{table:fcos_crowd_human}, \Ours compares favorably with its anchor-based counterpart RetinaNet on two out of three metrics (AP and JI), \textit{which suggests that anchor-based detectors have no large advantages even under the highly crowded scenario.} The higher MR$^{-2}$ of \Ours denotes that \Ours might have a large number of false positives with high confidence. By using the center-ness, MR$^{-2}$ can be significantly reduced from $59.04\%$ to $51.34\%$. As a result, \Ours\ can achieve better results under all the three metrics.

    Furthermore, as shown in \cite{chu2020detection}, it is more reasonable to let one proposal make multiple predictions under the highly crowded scenario (\ie, multiple instance prediction (MIP)). After that, these predictions are merged by Set NMS \cite{chu2020detection}, which skips the suppression for the boxes from the same location. A similar idea can be easily incorporated into \Ours. To be specific, if a location should be associated to multiple objects, instead of choosing a single target (\ie, the closest one to the location), the location's targets are set as the $K$-closest objects.
    Accordingly,
    the network is required to make $K$ predictions per location. Moreover, we do not make use of the earth mover's distance (EMD) loss for simplicity. Finally, the results are merged by Set NMS \cite{chu2020detection}. As shown in Table~\ref{table:fcos_crowd_human}, with MIP and Set NMS, improved performance is achieved under all the three metrics.

    \section{Conclusion}
    In this work, we have proposed an anchor-free and proposal-free one-stage detector \Ours.
    Our experiments demonstrate that
    \Ours\ compares favourably against the %
    widely-used
    anchor-based one-stage detectors, including RetinaNet, YOLO and SSD, but with much less design complexity. \Ours\ completely avoids all computation and hyper-parameters related to anchor boxes and solves the object detection in a per-pixel prediction fashion, similar to other dense prediction tasks such as semantic segmentation.  \Ours\ also achieves state-of-the-art performance among one-stage detectors. We also present some real-time models of our detector, which have state-of-the-art performance and inference speed. Given its effectiveness and efficiency, we hope that \Ours\ can serve as a strong and simple alternative of current mainstream anchor-based detectors.

    \section*{Acknowledgments}
    This work was in part supported by ARC DP grant \#DP200103797.
    We would like to thank the author of \cite{fcosplus} for %
    the tricks of
    center sampling and GIoU. We also thank Chaorui Deng
    for
    his suggestion of positioning the center-ness branch with box regression.

\balance
\bibliographystyle{ieeetr}
\bibliography{FCOS_TPAMI}

\begin{IEEEbiographynophoto}
{Zhi Tian}
 is a PhD student at School of Computer Science, The University of Adelaide,
 Australia. He received his B.E. degree in Computer Science and Technology from Sichuan University, China. He was awarded a Google PhD fellowship in 2019.
\end{IEEEbiographynophoto}

\begin{IEEEbiographynophoto}
{Chunhua Shen}
 is a Professor at School of Computer Science, The University of Adelaide,
 Australia.
\end{IEEEbiographynophoto}

\begin{IEEEbiographynophoto}
{Hao Chen}
 is a PhD student at School of Computer Science, The University of Adelaide,
 Australia.
 He received his BSc and master degrees from Zhejiang University, China.

\end{IEEEbiographynophoto}

\begin{IEEEbiographynophoto}
{Tong He}
 is a postdoc researcher at The University of Adelaide. He did his PhD study at the same university.
 He received his master degree
 from Wuhan University, China.
\end{IEEEbiographynophoto}

\end{document}